# ChatGPT-4 and Other LLMs in the Turing Test: A Critical Analysis

Marco Giunti
giunti@unica.it
Università di Cagliari
Dip. di Pedagogia, Psicologia, Filosofia

**Abstract**
This paper critically examines the recent publication “ChatGPT-4 in the Turing Test” by Restrepo Echavarría (2025), challenging its central claims regarding the absence of minimally serious test implementations and the conclusion that ChatGPT-4 fails the Turing Test. The analysis reveals that the criticisms based on rigid criteria and limited experimental data are not fully justified. More importantly, the paper makes several constructive contributions that enrich our understanding of Turing Test implementations. It demonstrates that two distinct formats, the three-player and two-player tests, are both valid, each with unique methodological implications. The work distinguishes between absolute criteria for passing the test—the machine’s probability of incorrect identification equals or exceeds the human’s probability of correct identification—and relative criteria—which measure how closely a machine’s performance approximates that of a human—, offering a more nuanced evaluation framework. Furthermore, the paper clarifies the probabilistic underpinnings of both test types by modeling them as Bernoulli experiments—correlated in the three-player version and uncorrelated in the two-player version. This formalization allows for a rigorous separation between the theoretical criteria for passing the test, defined in probabilistic terms, and the experimental data that require robust statistical methods for proper interpretation. In doing so, the paper not only refutes key aspects of the criticized study but also lays a solid foundation for future research on objective measures of how closely an AI’s behavior aligns with, or deviates from, that of a human.

## 1. Introduction

Recently the Turing Test has been the focus of renewed interest (Gonçalves 2022, Bayne and Williams 2023, Biever 2023), largely due to the fact that Large Language Models (LLMs) appear to have achieved a level of linguistic proficiency similar to that of humans, and exhibit remarkable abilities in logical reasoning, creative problem solving, contextual understanding, and synthesis of complex knowledge, thus reflecting multiple aspects of human cognitive abilities. In particular, several recent studies (Jannai et al. 2023; Jones and Bergen 2024a, 2024b, 2025; Jones et al. 2025; Restrepo Echavarría 2025) have reported on tests administered to ChatGPT-4 and other LLMs. These tests are more or less similar to the imitation game originally proposed by Alan Turing (1950) to determine whether a machine is capable of thought. This paper is specifically dedicated to an in-depth critical analysis of Restrepo Echavarría's article, published online in *Minds and Machines* on January 25, 2025.

This choice is motivated by two main reasons. First, the cited article maintains three very precise and bold theses:

(1) "No minimally serious implementation of the test has been reported to have been carried out" (p. 1).
(2) "This paper reports on a series of runs of a minimally valid version of the Turing Test with ChatGPT-4" (p. 2).
(3) Based on the test results, "it is safe to reject the hypothesis that ChatGPT-4 passes the Turing Test" (p. 5).

Second, the recognized authority of the journal in which the article was published implies that these theses should be taken seriously as potential cornerstones of the current debate on the Turing Test and LLMs.

The critical analysis undertaken in this paper, however, will demonstrate that none of the three theses is justified. More specifically, it will be shown that:

- Thesis (1) is unjustified because the tests reported in Jones and Bergen (2024a, 2024b, 2025) and Jones et al. (2025) cannot be dismissed as "not minimally serious implementations of the Turing Test" based on the five criteria proposed by Restrepo Echavarría (2025).
- There are good reasons to conclude that the test described in the cited work is not "a minimally valid version of the Turing Test".
- The test results do not allow us to reject the hypothesis that ChatGPT-4 passes the Turing Test.

Moreover, the analysis establishes several positive contributions that clarify various controversial points regarding the structure and interpretation of the Turing Test.

First, it is shown that there exist at least two equally valid ways to implement the test: the three-player version and the two-player version.

Second, it is explained how the difference between the two types of tests implies that the criteria for passing the test differ, yet they are two specific formulations of the same more general criterion. It is also necessary to distinguish between absolute criteria, which define the machine's optimal performance, and relative criteria, which establish how closely its performance approximates that optimum.

Third, the criteria for passing the test are formulated in probabilistic terms, but this presupposes that the probabilistic structure of the test is explicitly defined beforehand for both the three-player and the two-player versions. These formal definitions are provided in the Appendix.

Fourth, a clear distinction is made between the theoretical criteria for passing the test and the experimental results. The former are expressed in terms of probabilities, while the latter merely record the number of trials along with the percentages of correct and incorrect identifications. Experimental results, on their own, are insufficient to draw definitive conclusions about whether the test has been passed or failed; rather, such conclusions can only be reached by applying appropriate statistical methods to the data.

## 2. Have There Been Any Minimally Serious Implementations of the Turing Test?

The justification for thesis (1)—no such implementations exist—rests on five criteria that, according to the author, are "beyond reasonable doubt, essential features any valid Turing Test will have" (Restrepo Echavarría 2025, p. 3). In summary, these essential features are:

(i) "It is a test that involves three entities: an interrogator, another human and a machine. Two entity versions are not valid, as the need for comparison by the interrogator is essential." (p. 3)

(ii) "The interrogator knows that s/he is talking to another person and a machine, and has to identify who is who through open-ended conversation. [...] The other person and the machine also have the information of what the game is about and their role in trying make the interrogator identify them as the person." (p. 3)

(iii) "The machine passes the Turing Test if the interrogator cannot reliably tell who is who. Conversely, the machine does not pass the Turing Test if the interrogator can reliably tell (significantly more than chance) who is who. Turing's standard for passing the Turing Test has been confused with his prediction that sometime around the year 2000 there may well be a computer able to fool judges 30% of the time, one of these being the case of Eugene Goostman." (p. 3)

(iv) "The interactions need to be of non-trivial time. [...] Reasonable and practicable time has to be given for participants to be able to think and write their lines. [...] Nothing shorter than five minutes should be considered minimally serious." (p. 3)

(v) "Off-hand speculations by experts that machine X would pass the test are not relevant. [...] Preliminary speculations and interactions are fine as 'giggle tests'. But these are not runs of the Turing test and cannot claim to be such." (p. 3)

Among these five criteria, those most aligned with Turing's original formulation of the imitation game, and with the widely shared subsequent interpretations of the test, are criteria (ii) and (v). By contrast, criteria (i), (iii), and (iv) need further discussion and support to be taken as essential elements of any valid Turing Test.

### *2.1 Discussion of Criterion (i)*

In particular, regarding (i), it is true that the test originally proposed by Turing was a three-player test, but this is not the form in which the test has been—and still is—usually implemented. Speaking of the first attempts to actually implement the test, Saygin, Cikekli, and Akman noted:

> The TT has never been carried out in exactly the same way Turing originally described. However, there are variants of the original in which computer programs participate and show their skills in 'humanness'. (Saygin et al. 2000, p. 501)

At present, the two-player version is implemented more frequently (Saygin et al. 2000, p. 467; French 2000, pp. 2–3; Jannai et al. 2023; Jones and Bergen 2024a, 2024b; Jones et al. 2025). However, departing from this trend, Jones and Bergen (2025) report on the three-player test they administered to three different LLMs and to a version of Eliza.

In the two player test, a human interrogator converses with a single respondent, who may be either a human or a machine. This version of the test cannot be dismissed simply because "the need for comparison by the interrogator is essential" (Restrepo Echavarría 2025, p. 3). In fact, if criterion (ii) is met, the two-player test also implies a comparison by the interrogator between the responses received and those expected from a human or a machine.

Furthermore, it is important to note that two-player implementations typically involve several interrogations, some with a machine respondent and others with a human. This guarantees a comparison between the machine's passing rate and that of the human, i.e., between the machine's percentage of incorrect identifications and the human's percentage of correct identifications.

### 2.2 Discussion of Criterion (iii)

With regard to criterion (iii), it is first necessary to specify more precisely what it means that "the interrogator cannot reliably tell who is who." Restrepo Echavarría clarifies (p. 4) that this means that, at the end of the conversation, the interrogator must have a 50% chance of making an incorrect (or correct) identification. However, the author did not question why the 50% probability is not usually chosen as an adequate threshold for declaring that the test has been passed.

The point is that this probability is the threshold for machine's optimal performance in a three player test. This means that the machine passes the test if the probability of incorrect identification is equal to or greater than 50%. In the three-player test, a 50% probability of incorrect identification is the minimum requirement for passing, which is equivalent to saying that a 50% probability of correct identification is the maximum requirement. However, establishing merely this *absolute* threshold does not allow us to appreciate how closely the machine's performance approaches that of a human. Turing himself speculated that a probability of incorrect identification of at least $30\%$ would be a reachable goal in about fifty years,[1] indicating acceptable, though not optimal, performance by the machine.[2] Although there is some arbitrariness in setting a threshold for acceptable performance, Turing's choice is less arbitrary than it may initially seem. In fact, if we compare the 30% probability to the optimal threshold of $50\%$, we obtain $30/50 = 6/10$. In other words, a 30% probability corresponds to $6/10$ of the optimal threshold and therefore, in agreement with the interpretation of Saygin, Cikekli, and Akman (2000, p. 501) and French (2000, p. 3)[3], we can affirm that the machine demonstrates, in the test, at least a sufficient "degree of humanness."

Moreover, we must also consider that an acceptable criterion for passing the test must be general, in the sense that it must be uniformly applicable to the different ways in which the test can be conducted. The two-player implementation of the Turing Test is currently the most common, so it is reasonable to question whether the 50% probability of incorrect (or correct) identification has the same significance in this test as in the three-player test.

---

[1] The original formulation of Turing's hypothesis (1950, p. 442) is equivalent to this, but it considers the probability of correct identification, not the probability of incorrect identification: "I believe that in about fifty years' time it will be possible to programme computers [...] to make them play the imitation game so well that an average interrogator will not have more than 70 per cent. chance of making the right identification after five minutes of questioning."

[2] Warwick and Shaw (2016) claimed that this milestone was achieved by Eugene Goostman, a program that was identified as human 10 times out of 30 (33%) in a Turing test held at the Royal Society in London on 6 and 7 June 2014. In fact, Warwick and Shaw's thesis is not justified. For details, see Sec. 4 in the Appendix.

[3] "I believe that in 300 years' time people will still be discussing the arguments raised by Turing in his paper. It could even be argued that the Turing Test will take on an even greater significance several centuries in the future when it might serve as a moral yardstick in a world where machines will move around much as we do, will use natural language, and will interact with humans in ways that are almost inconceivable today. In short, one of the questions facing future generations may well be, 'To what extent do machines have to act like humans before it becomes immoral to damage or destroy them?' And the very essence of the Turing Test is our judgment of how well machines act like humans." (French 2000, p. 3)

We have seen above that, in the three-player test, a 50% probability corresponds to a machine's optimal performance because it is achieved when, at the end of the test, both the machine's and the human's answers attest the same "degree of humanness" of the two respondents. Moreover, it should be noted that in the three-player test the only two possible outcomes are an incorrect identification or a correct one for *both* respondents. This implies that the machine's probability of incorrect identification and the human's probability of incorrect identification are always equal. But since the probability of incorrect identification is equal to one minus the probability of correct identification, this also implies that the machine's probability of incorrect identification is equal to the human's probability of correct identification if, and only if, both are 50% (see Sec. 1 in the Appendix for a formal proof).

On the contrary, none of this occurs in the two-player test. In this type of test, the machine and the human are interrogated separately and, therefore, at the end of the two interrogations, the responses of the machine and the human determine their respective probabilities of incorrect identification, but these are not necessarily equal (see Sec. 2 in the Appendix for formal details). In particular, suppose that the machine's probability of incorrect identification is 50%, but the human's probability of incorrect identification is lower—say, 25%—or, equivalently, the human's probability of correct identification is 75%. It is evident that, under these assumptions, the machine's performance would not be optimal, because to match that of the human, the machine's probability of *incorrect* identification would need to be increased to at least 75%, that is, it would have to equal or exceed the human's probability of *correct* identification.

We can therefore conclude that the criterion proposed for passing the test—50% probability—is not adequate for the following reasons:

(I) It serves only as an absolute criterion, defining the threshold for machine's optimal performance, but does not measure how closely its performance approximates that of a human.

(II) As an absolute criterion, it is valid only for the three-player test, not generally.

The discussion above, however, also leads to two positive conclusions:

(III) If we are interested in an absolute criterion for passing the test, one that applies equally to both the three-player and the two-player test, we must require that the machine's probability of incorrect identification equals or exceeds the human's probability of correct identification. In the three-player test, this criterion reduces to the 50% threshold, because we have seen that the two probabilities in question are equal if and only if both are 50%.[4] In the two-player test, however, it involves comparing the two probabilities obtained at the end of the two separate interrogations of the machine and the human. Nonetheless, the stated criterion establishes the machine's optimal performance in the same sense in both the three-player and two-player tests, as it requires the machine to have at least the same probability as a human of being recognized as human.

(IV) If we are also interested in evaluating to what extent the machine's performance approaches optimal performance, we must consider the ratio between the machine's probability of incorrect identification and the probability threshold for machine's optimal performance. Note that, in the two-player test, the threshold is equal to the human's probability of correct identification, whereas in the three-player test it is equal to $50\%$,[5] as explained above. Regarding the three-player test, we have already applied this criterion to the case hypothesized by Turing—a $30\%$ probability of incorrect identification—yielding $30/50 = 0.6$. Instead, as an example for the two-player test, we can consider the case hypothesized above, namely: a machine's probability of incorrect identification of 50% and a human's probability of correct identification of $75\%$; hence, we obtain $50/75 = 0,\overline{66}$.

[4] Jones and Bergen (2025, pp. 4-5) report that GPT-4.5, equipped with an appropriate persona-prompt, achieved a statistically significant win rate > 50% in two different three-player tests. If their results are confirmed, they constitute the first experimental demonstration that a machine passed a Turing test in accordance with the absolute passing criterion defined here.

### *2.3 Discussion of Criterion (iv)*

The first part of criterion (iv) is fully acceptable, but the second part—stating "Nothing shorter than five minutes should be considered minimally serious." (p. 3)—requires further discussion.

Firstly, setting a reasonable time limit for the test is a practical necessity to facilitate its implementation. In theory, the test duration should be open-ended, allowing the interrogator to decide when to conclude the conversation and deliver a verdict. In practice, however, this would complicate the implementation, or even render the test unfeasible.

Turing himself, when envisioning an actual test, mentioned "5 minutes of interrogation" (1950, p. 442). It is important to note, however, that this limit was intended by Turing to be a *maximum* duration rather than a *minimum*, as required by criterion (iv) instead. No justification has been provided to support this significant reversal of interpretation. Moreover, the five-minute maximum has traditionally been considered adequate for the design of an effective test implementation, and it has been adopted in most tests conducted in the past, including the most recent ones involving ChatGPT-4 (Jones and Bergen 2024a, 2024b; Jones et al. 2025) or other LLMs (Jones and Bergen 2025).

Due to the lack of adequate argumentation supporting the new interpretation of a minimum five-minute duration, the second part of criterion (iv) is not acceptable.

### *2.4 Summing Up*

As mentioned in the introduction, apart from the discussed article, at least five other recent studies (Jannai et al. 2023; Jones and Bergen 2024a, 2024b, 2025; Jones et al. 2025) describe different versions of the Turing Test administered to ChatGPT-4 and other LLMs. However, as these works reveal, none of these tests satisfies all five criteria proposed by Restrepo Echavarría (2025). Although one might agree with the author that no study in the literature describes an implementation that meets all these criteria (Oppy and Dowe 2021), one must ask whether, on this basis, thesis (1) is justified:

> There has been considerable optimistic speculation on how well ChatGPT-4 would perform in a Turing Test. However, *no minimally serious implementation of the test has been reported to have been carried out*. (p. 1, italics mine)

Our discussion has established that criteria (i) and (iii) are not acceptable, nor is the second part of criterion (iv). The question then arises whether the remaining criteria—(ii) and (v), along with the first part of (iv)—are sufficient to exclude the more recent studies describing various Turing Test versions administered to ChatGPT-4 and other LLMs.

---

[5] The different thresholds in the two types of test reflect their different structures. In the two-player test, the questioning of the human and the questioning of the machine are carried out in such a way that the identification of one is independent of the identification of the other. This is never the case in the three-player test because, due to the way the three-player test is constructed, the identification of the human is necessarily the opposite of the identification of the machine.

Regarding criterion (v), it is clearly satisfied by all the tests described in those works, since none of these tests falls within the types of hypotheses, speculations, or procedures excluded by criterion (v).

As for criterion (ii), it is not satisfied by the test described in Jannai et al. (2023). First of all, it should be noted that this is a two-player test in which the roles of the interrogator and the respondent, as well as their respective tasks, are not clearly distinguished. In fact, when both interlocutors are human, the explicit task assigned to them is identical—namely, to determine whether the other is a human or a machine. Moreover, the two participants are free to add additional motivations, such as pretending to be an AI, convincing the other that they are conversing with a human, etc. (Jannai et al. 2023, p. 2). Instead, the tests described in Jones and Bergen (2024a, 2024b) and Jones et al. (2025) fully satisfy criterion (ii). These are also two-player tests, but the roles of the interrogator and the respondent (referred to as the "witness") are clearly distinguished and their respective tasks are explicitly assigned (Jones and Bergen 2024a, p. 5186; Jones and Bergen 2024b, pp. 3 and 8; Jones et al. 2025, p. 1618). Also the three-player tests described in Jones and Bergen (2025, pp. 2 and 4, Figure 6) satisfy criterion (ii).

Finally, regarding the first part of criterion (iv), it is not satisfied by the test described in Jannai et al. (2023), because that test has a maximum duration of only 2 minutes and also imposes a maximum limit of 100 characters and 20 seconds for each message (Jannai et al. 2023, p. 3). These very strict limits do not allow for the formulation of sufficiently complex questions and answers, nor for the in-depth discussion of a conversational topic, thereby violating the first part of criterion (iv). Instead, the tests described in Jones and Bergen (2024a, 2024b, 2025) and Jones et al. (2025) satisfy the first part of criterion (iv). In those tests, the maximum duration is the traditional 5 minutes, the maximum character limit per message is raised to 300, and within the 5-minute time window no maximum time limit is set for each message (Jones and Bergen 2024a, pp. 5185–6; Jones and Bergen 2024b, p. 8; Jones and Bergen 2025, pp. 13-14; Jones et al. 2025, p. 1618).

In conclusion, we have seen that the only justification proposed for thesis (1) is the observation that there are no studies in the literature describing an implementation of the Turing Test that meets all the criteria (i)–(v). However, of these five criteria, we have established that (i), (iii), and the second part of (iv) are not acceptable. We then asked whether the remaining criteria are sufficient to exclude the more recent works that describe various versions of the Turing Test administered to ChatGPT-4 and other LLMs. The answer was positive for Jannai et al. (2023), but negative for Jones and Bergen (2024a, 2024b, 2025) and Jones et al. (2025). It follows, therefore, that thesis (1) is unjustified.

## 3. Is Restrepo Echavarría's Test Minimally Valid?

Thesis (2) asserts that the work in question "reports on a series of runs of a minimally valid version of the Turing Test with ChatGPT-4" (p. 2). However, the detailed description of the test's structure and administration occupies only the second paragraph (11 lines) of section 2. The first paragraph introduces that description with a merely rhetorical question and the remainder of section 2 is devoted to discussing the test results, which I will analyze in detail later.

From this brief description, we can extract the following data: (a) The test was conducted as a three-player version, since in each trial a person acted as the interrogator and, as respondents, another person and ChatGPT-4, whose identities were to be determined by the interrogator at the end of the trial; (b) each trial comprised two successive conversations—one with ChatGPT-4 and the other with the human respondent—with the order of these chats (conducted via WhatsApp) varying from trial to trial; (c) in total, there were 10 trials involving 10 participants (5 teachers and 5 undergraduate students), none of whom had specialized AI knowledge; (d) each person assumed the role of interrogator only once and, presumably, the role of respondent only once; (e) in all 10 trials, a single prompt was used for ChatGPT-4, instructing it that it was participating in a Turing Test and advising it on behaviors intended to avoid immediately revealing its identity; (f) regarding duration, the only reported datum is: "In 9 out of 10 runs, the interrogators correctly identified the human and the machine after a 7-minute chat." However, it remains unclear whether the 7 minutes refer to each of the two separate chats in a trial or to the trial as a whole.

Based on the discussion in the previous section, we cannot rely on criteria (i)–(v) to determine whether this version of the Turing Test is minimally valid. Nonetheless, at least two aspects of the experimental design cast serious doubts on its validity.

First, the number of trials conducted (10) is small and certainly not sufficient to draw conclusions that have robust statistical value. Even the test conducted at the Royal Society of London in 2014, although questionable in many respects, included 30 trials for each competing program (Warwick and Shaw 2016). If we then consider the other tests that ChatGPT-4 and other LLMs have recently been subjected to, the number of trials on which they are based are of other orders of magnitude: several thousand conversations in total for the different tests described in Jones and Bergen (2024a, p. 5186; 2024b, p. 8; 2025, p. 14) and Jones et al. (2025, p. 1618), and even more than 10,000,000 conversations for Human or Not (Jannai et al. 2023, p. 2).

Second, the test described above was conducted using only a single prompt for ChatGPT-4. However, it is well known that one of the most significant characteristics of LLMs is their ability to dramatically alter their behavior depending on the prompt provided. Given this property, the fact that a single ChatGPT-4 prompt fails the test does not justify concluding that *the model* is incapable of passing it. Rather, the inadequacy of *the prompt* should be considered, at least in the first instance. From this point of view, the five studies cited above demonstrate a deeper awareness of the problem. Jannai et al. (2023) employed a wide range of prompts—including variations in personal background, personality traits, relevant information, as well as writing and language styles—while Jones, Bergen and collaborators created and experimented with 45 different prompts (Jones and Bergen 2024a, p. 5185), from which the best was selected and further tested in statistically controlled experiment involving several different pre-registered hypotheses (Jones and Bergen 2024b, pp. 2–3, 7, 9; Jones et al. 2025, p. 1617; Jones and Bergen 2025, pp. 2, 12, 15-16).

To be fair, it must be mentioned that in section 3, Restrepo Echavarría reports having subsequently modified the prompt, without, however, obtaining any significant improvements in performance in other interviews conducted by the interrogators. Nonetheless, these observations are not substantiated by the description of any systematic experimentation on the subject and therefore fall precisely into that category of "off-hand speculations" or "preliminary speculations and interactions" that the author himself has deemed unacceptable according to his criterion (v).

## 4. Do the Test Results Allow Us to Reject the Hypothesis That ChatGPT-4 Passes the Turing Test?

As previously reported, the result of the test conducted by Restrepo Echavarría was that, in 9 out of 10 trials, the interrogators correctly identified the human and the machine after a 7-minute conversation. Based on this result, the author presents a probabilistic argument intended to justify thesis (3), namely that "it is safe to reject the hypothesis that ChatGPT-4 passes the Turing Test" (p. 5). The argument is structured as follows:

1. If ChatGPT-4 passes the Turing Test, then the probability $p$ of being correctly identified in each trial is 50%.
2. If $p = 50\%$ in each trial, then we can use the binomial distribution formula to compute the probability of obtaining $k = 9$ correct identifications out of $n = 10$ trials; i.e., by indicating such event with $E_k^n$:
$$\mathbf{P}(E_k^n) = \binom{n}{k} p^k (1-p)^{n-k}$$
3. Carrying out the calculations yields:
$$\mathbf{P}(E_k^n) = 5/512 = 0{,}009765625$$
4. Thus, if $p = 50\%$ in each trial, the probability of obtaining 9 correct identifications out of 10 trials is approximately $0{,}98\%$.
5. "That is, the probability that it is an error to say that the machine does not pass the Turing Test, in light of these results, is less than $1\%$." (p. 4)
6. "Consequently, it is safe to reject the hypothesis that ChatGPT-4 passes the Turing Test." (p. 5)

Step 1 of the argument states that passing the Turing Test implies that the probability of correct identification in each trial is 50%. But we have seen that 50% of correct identification is the maximum probability threshold for machine's optimal performance in a three-player test. Therefore, it is a sufficient condition for passing the test, but not a necessary one, as any other probability below threshold will do. Therefore, premise 1 is not justified.

Step 2 asserts that the binomial distribution formula can be used to compute $\mathbf{P}(E_k^n)$, if $p = 50\%$ in each trial. This step is justified, as all three conditions needed for applying the binomial distribution formula are met. Specifically: (i) each repetition of the test is a Bernoulli experiment (see Secs. 1, 2, 3 in the Appendix), (ii) the antecedent of the implication in step 2 asserts that each repetition has the same probability $p = 50\%$ and, presumably, (iii) the 10 trials were independent.

Step 3 is justified, as it follows from step 2 by purely mathematical derivations.

Step 4 is also justified, as it follows from steps 2 and 3. Thus, only the last two steps remain to be analyzed.

Step 5 asserts that the probability that ChatGPT-4 passes the Turing Test, "in the light of these results", is less than $1\%$. Let us denote by $\mathcal{A}$ the assertion: "The probability that ChatGPT-4 passes the Turing Test is less than $1\%$." $\mathcal{A}$ is thus presented as an obvious consequence of the implication obtained at step 4 and the observed result of 9 correct identifications out of 10 trials. However, the inference of $\mathcal{A}$ from these two premises is clearly invalid from a deductive standpoint. What remains to be assessed is whether $\mathcal{A}$ can be derived by applying appropriate statistical methods.

Step 6 states that the hypothesis that ChatGPT-4 passes the Turing Test can be safely rejected. If the previous step had indeed established that this hypothesis has a probability of less than 1%, this conclusion would be justified.

Let us now ask what we can conclude, from a statistical perspective, based on the result of 9 correct identifications out of 10 trials. First, we note that the experiment conducted consists of a sequence of $n = 10$ Bernoulli *trials*, as conditions (i), (ii), and (iii) are satisfied (see the discussion of step 2). In general, counting the number $k$ of successes observed in such an experiment serves to test the hypothesis that the probability $p$ of each trial has the hypothesized value $p_0$. In this case, the hypothesis to be tested is $p = 0.5$, which, as we have seen, represents the maximum probability of correct identifications for passing the three-player Turing Test.

As is well known, the number of successes $k$ in $n$ trials (which we denote as $E_k^n$) has no intrinsic statistical significance. However, once a given level of statistical significance has been established, we can reject the hypothesis $p = p_0$ if and only if the result $E_k^n$ is statistically significant. We denote the statistical significance level as $\alpha$. In the social sciences, at least $\alpha = 5\%$ is typically used, and the more stringent $\alpha = 1\%$ is not uncommon. As suggested in step 5 of the argument in question, we choose $\alpha = 1\%$. Furthermore, recall that $E_k^n$ is statistically significant at level $\alpha = 1\%$ if and only if the sum of the probability of $E_k^n$ and those of all equally or less probable results[6] is less than $1\%$.

Let us now determine whether $E_9^{10}$ is statistically significant at the 1% level. As can be easily verified, it is not because the sum of $\mathbf{P}(E_9^{10})$ and the probabilities of all equally or less probable results, namely $\mathbf{P}(E_1^{10})$, $\mathbf{P}(E_{10}^{10})$, and $\mathbf{P}(E_0^{10})$, is:

$$\mathbf{P}(E_9^{10}) + \mathbf{P}(E_1^{10}) + \mathbf{P}(E_{10}^{10}) + \mathbf{P}(E_0^{10}) = 0.009765625 + 0.009765625 + 0.000976563 + 0.000976563 = 0.021484376 > 0{,}01$$

Consequently, we cannot statistically reject the hypothesis $p = 0.5$ and, *a fortiori*, that ChatGPT-4 passes the Turing Test, since rejecting the latter is equivalent to rejecting *all* hypotheses $p = p_0$, with $p_0 \leq 0.5$. However, this does not mean that any of these is confirmed. Since the result is known, we can in fact determine all values of $p$ that are *compatible* with this result at the $\alpha = 1\%$ level. Rounding to the second decimal place, we obtain the closed interval $[0.49, 0.99]$. Indeed, for all values of $p$ within this interval, $E_9^{10}$ is not statistically significant at the $1\%$ level. This means that for any $p \in [0.49,\ 0.99]$, $E_9^{10}$ belongs to the smallest set of more probable results whose probabilities, added together, equal or exceed 99%.[7] For all values of $p \in [0, 0.48]$ and $p = 1$, $E_9^{10}$ is instead statistically significant at the $1\%$ level.[8]

For completeness, let us also consider what happens if, instead of $\alpha = 1\%$, we use the statistical significance level $\alpha = 5\%$. With this higher significance level $E_9^{10}$ becomes statistically significant, and thus, under this more permissive statistical convention, we can reject the hypothesis $p = 0.5$. Finally, we estimate the range of values of $p$ compatible with the result $E_9^{10}$ at the 5% level. Rounding to the second decimal place, we obtain $p \in [0.56, 0.99]$. For all values of $p \in [0, 0.55]$ and $p = 1$, $E_9^{10}$ is instead statistically significant at the 5% level, and we can thus reject the hypothesis that ChatGPT-4 passes the Turing Test.

To summarize, we have seen that, based on the result of 9 correct identifications out of 10, it is possible to reject the hypothesis that ChatGPT-4 passes the Turing Test only by assuming the significance level $\alpha = 5\%$. However, even with this higher level, the range of probabilities compatible with the result remains very broad. Moreover, we must remember that these are probabilities of correct identifications. This means that the probabilities of incorrect identifications compatible with the result fall within the interval $[0.01, 0.44]$. Therefore, even conceding that the test has demonstrated that ChatGPT-4 has not passed the Turing Test in absolute terms, we can say very little about its relative performance or "degree of humanness", which remains compatible with values ranging from $1/50 = 2\%$ to $44/50 = 88\%$, i.e., from nearly total failure to very good performance. Finally, if we consider the significance level $\alpha = 1\%$ instead, the test proves to be even less informative, since the "degree of humanness" demonstrated by ChatGPT-4 is now compatible with any value between $1/50 = 2\%$ and even $51/50 = 102\%$.

---

[6] Such a sum is known as the *two-tailed exact p-value* of the result $E_k^n$ under the hypothesis $p = p_0$.

[7] This confidence interval, known as *Sterne's exact central interval* (Sterne 1954), is computed as follows. Given the observed result $E_k^n$, we take the central value $p_0 = k/n$, rounded to the $d$-th decimal place. Starting from this value, we increase $p_0$ in steps of $10^{-d}$: for each $p_0$, we compute the corresponding two-tailed exact p-value. The right bound of the interval is defined as the largest $p_0$ whose p-value is greater than or equal to the significance level $\alpha$; the left bound is obtained analogously by decreasing $p_0$ in steps of $10^{-d}$.

[8] This means that, for all $p \in [0, 0.48]$ and $p = 1$, the sum of the probability of $E_9^{10}$ and those of all equally or less probable results is less than 1% or, equivalently, $E_9^{10}$ belongs to the larger set of less probable results whose probabilities, added together, are less than 1%. Because of the approximation to the second decimal place, the statistical significance or non-significance of $E_9^{10}$ is not determined for any value of $p$ belonging to the open intervals $(0.48, 0.49)$ $and$ $(0.99, 1)$. However, the width of these two intervals can be made as small as desired by choosing a finer approximation grid.

## 5. Conclusions

The analysis conducted so far has established that none of the three theses (1), (2), or (3) is justified. However, beyond this negative conclusion, the analysis has also led to a series of positive findings.

First, we have highlighted that there are at least two different ways in which the Turing Test can be effectively implemented: as a three-player or a two-player test. The first type of implementation adheres more closely to Turing's original formulation, while the second is usually recognized as the standard implementation. However, both versions are well represented by the most recent and statistically robust tests to which ChatGPT-4 and other LLMs have been subjected (Jones and Bergen 2024a, 2024b, 2025; Jones et al. 2025). All of them satisfy those criteria for minimal validity that our critical analysis has found acceptable.

Second, regarding the criterion for passing the test, we have seen that it is essential to distinguish between an absolute criterion and a relative criterion. In the case of the three-player test, the absolute criterion is represented by at least $50\%$ probability of incorrect identification or, equivalently, at most $50\%$ probability of correct identification. In the two-player test, however, the machine's probability of incorrect identification must equal or exceed the human's probability of correct identification. The relative criterion provides a measure of how much the machine's performance approaches, or even exceeds, the threshold for optimal performance. In the case of the three-player test, it is given by the ratio between the actual probability of incorrect identification and the 50% probability threshold. For the two-player test, the relevant measure is the ratio between the machine's probability of incorrect identification and the human's probability of correct identification.

Third, the criteria proposed for passing the Turing Test presuppose the concept of probability and therefore require that the probabilistic structure of the test be explicitly defined in advance. In the Appendix, we have shown that, for both the three-player and two-player versions, this structure consists of two Bernoulli experiments, which differ only in that, in the three-player test, they are logically correlated, whereas in the two-player test, they are not. This implies that, in the three-player test, the equality between the machine's probability of incorrect identification and the human's probability of correct identification occurs if and only if both probabilities are 50%. In contrast, in the two-player test, the value at which this equality occurs cannot be determined a priori.

Fourth, it is important to distinguish between the criteria for passing the test, which are theoretical criteria formulated in terms of appropriate probabilities, and the experimental results, which report the number of trials and the percentages of correct or incorrect identifications, both for the machine and for the human. These results, by themselves, cannot determine whether the test has been passed or failed, either in absolute or relative terms.

Instead, this can only be established by applying appropriate statistical methods that allow experimental results to be correlated with theoretical criteria.

Finally, regarding the meaning of the test, we have seen that passing it implies demonstrating a “degree of humanness” at least equal to that of a human. This is the most straightforward interpretation and the one most consistent with the test's structure. However it differs from Turing’s original interpretation, in which the test would serve as a sufficient criterion for determining a machine’s intelligence, or even its capacity for thought. Nevertheless, we believe that this minimal interpretation does not diminish the test's value. Rather, it highlights the test's current and future significance. Nowadays it has become crucial to establish objective criteria indicating how closely an AI’s behavior aligns with—or deviates from—that of a human. It is also plausible that evaluations of this kind will guide us in the inevitable integration of AIs into our social and cultural lives in the future.

## Appendix

In order to establish a criterion for passing the Turing Test—whether interpreted as a three-player or a two-player test—it is necessary to employ the concept of probability. We must therefore first clarify the random experiments involved in each version of the test, together with the corresponding sample spaces and associated sigma-algebras, on which the relevant probabilities are defined.

### *1. Formalization of the Three-Player Test as Two Logically Correlated Bernoulli Experiments*

Consider first the three-player test and ask what the probabilistic structure of a single test is. In fact, the three-player test exhibits a rather complex probabilistic structure, and the best way to capture this complexity is to view a single three-player test as the execution of two logically correlated random experiments—each of which has only two possible outcomes (i.e., it is a Bernoulli experiment). We elaborate this point below.

The two random experiments constituting the three-player test are the identifications of the two respondents by the interrogator. Denote the machine respondent as $m$, the human respondent as $h$, and the corresponding random experiments as $t_m$ and $t_h$. The identification of each respondent consists of asserting whether that respondent is a human or a machine. Let $M$ and $H$ be the two predicates corresponding to these assertions. Thus, the two possible outcomes of $t_m$ are $M(m)$ and $H(m)$, while those of $t_h$ are $M(h)$ and $H(h)$. In other words, the sample spaces for $t_m$ and $t_h$ are, respectively:

$$\Omega_m \coloneqq \{M(m), H(m)\}$$
$$\Omega_h \coloneqq \{M(h), H(h)\}$$

Since both sample spaces are discrete (indeed, finite), the associated sigma-algebras are simply their power sets $\mathcal{P}(\Omega_m)$ and $\mathcal{P}(\Omega_h)$. The four events of interest are the correct and incorrect identifications of the machine and the correct and incorrect identifications of the human. The former two are elementary events in $\mathcal{P}(\Omega_m)$, and the latter two are elementary events in $\mathcal{P}(\Omega_h)$:

$$\text{incorrect identification of the machine} = S_m \coloneqq \{H(m)\} \in \mathcal{P}(\Omega_m)$$
$$\text{correct identification of the machine} = F_m \coloneqq \{M(m)\} \in \mathcal{P}(\Omega_m)$$

$$\text{incorrect identification of the human} = F_h \coloneqq \{M(h)\} \in \mathcal{P}(\Omega_h)$$
$$\text{correct identification of the human} = S_h \coloneqq \{H(h)\} \in \mathcal{P}(\Omega_h)$$

Recall that, for any random experiment $t$ with sample space $\Omega$ and sigma-algebra $\Sigma(\Omega)$, an event $E \in \Sigma(\Omega)$ is said to *occur* $:\Leftrightarrow r(t) \in E$, where $r(t) \in \Omega$ is the actual outcome of $t$.

Denote the actual outcomes of $t_m$ and $t_h$ with $r(t_m) \in \Omega_m$ and $r(t_h) \in \Omega_h$, respectively, and let the probability functions on $\mathcal{P}(\Omega_m)$ and $\mathcal{P}(\Omega_h)$ be $\mathbf{p}_m$ and $\mathbf{p}_h$. Because the interrogator interacts with two respondents, $m$ and $h$, and can only provide two responses—$H(m) \wedge M(h)$ or $M(m) \wedge H(h)$—the two random experiments are logically correlated, so that we can establish a priori the following: $r(t_m) = H(m) \Leftrightarrow r(t_h) = M(h)$ and $r(t_m) = M(m) \Leftrightarrow r(t_h) = H(h)$. Thus, $S_m$ occurs $\Leftrightarrow F_h$ occurs and $F_m$ occurs $\Leftrightarrow S_h$ occurs. By the principle of equivalence,[9] it follows $\mathbf{p}_m(S_m) = \mathbf{p}_h(F_h)$ and $\mathbf{p}_m(F_m) = \mathbf{p}_h(S_h)$. Furthermore, since $F_m$ is the negation of $S_m$ and $F_h$ is the negation of $S_h$, the following holds: $\mathbf{p}_m(F_m) = 1 - \mathbf{p}_m(S_m)$ and $\mathbf{p}_h(F_h) = 1 - \mathbf{p}_h(S_h)$. Hence, $\mathbf{p}_m(S_m) = 1 - \mathbf{p}_h(S_h)$ and $\mathbf{p}_m(F_m) = 1 - \mathbf{p}_h(F_h)$. Therefore, by the first equality, $\mathbf{p}_m(S_m) = \mathbf{p}_h(S_h) \Leftrightarrow \mathbf{p}_m(S_m) = 0{,}5 = \mathbf{p}_h(S_h)$ and, by the second, $\mathbf{p}_m(F_m) = \mathbf{p}_h(F_h) \Leftrightarrow \mathbf{p}_m(F_m) = 0{,}5 = \mathbf{p}_h(F_h)$.

### *2. Formalization of the Two-Player Test as Two Uncorrelated Bernoulli Experiments*

In this version, the interrogator interacts with only one respondent at a time and knows that the respondent is either a machine or a human. The test is repeated several times, some with a machine and others with a human. Thus, this test is also formally represented by the two Bernoulli experiments specified earlier: $t_m$ when the respondent is a machine, and $t_h$ when the respondent is a human. The formalization of the two Bernoulli experiments is identical to that in the previous section. However, in this case, the experiments are not logically correlated. Therefore, it is not possible to derive either the equality between the machine's probability of incorrect identification $\mathbf{p}_m(S_m)$ and the human's probability of incorrect identification $\mathbf{p}_h(F_h)$, or the equality between the machine's probability of incorrect identification $\mathbf{p}_m(F_m)$ and the human's probability of correct identification $\mathbf{p}_h(S_h)$. The details are below.

Note that, by the way the two-player test is constructed, in each run the interrogator interacts with a single respondent, of which s/he knows only that it is either a machine or a human. Therefore, the two random experiments are not logically correlated; that is, we cannot a priori establish any of the following implications: $r(t_m) = H(m) \Rightarrow r(t_h) = M(h)$, $r(t_h) = M(h) \Rightarrow r(t_m) = H(m)$, $r(t_m) = M(m) \Rightarrow r(t_h) = H(h)$, $r(t_h) = H(h) \Rightarrow r(t_m) = M(m)$. Consequently, we cannot assert either that $S_m$ occurs $\Leftrightarrow F_h$ occurs or that $F_m$ occurs $\Leftrightarrow S_h$ occurs. Thus, unlike the case of the three-player test, the principle of equivalence is not applicable, and we cannot conclude that $\mathbf{p}_m(S_m) = \mathbf{p}_h(F_h)$ nor that $\mathbf{p}_m(F_m) = \mathbf{p}_h(S_h)$.

### *3. Alternative Formalization of the Three-Player Test as a Single Bernoulli Experiment*

An alternative way of formally representing the three-player test is to think of it as a single random experiment, rather than as two logically correlated random experiments (as was done above). From this point of view, the only random experiment constitutive of a single three-player test consists of the identification of both respondents by the interrogator. Again, this random experiment has only two possible outcomes and is thus a Bernoulli experiment. Moreover, as will be shown below, this formalization is equivalent to the previous one. However, it does not clearly capture the relationship between the probabilistic structure of the three-player test and that of the two-player test. On the contrary, as we have seen above, this relationship is evident under the previous formalization.

Similarly as before, we denote the machine respondent by $m$ and the human respondent by $h$, but the one random experiment by $t_{m,h}$. The identification of each respondent is the assertion that this respondent is either a human or a machine. We denote by $M$ and $H$ the two predicates corresponding to these assertions. Since one of the two respondents is a machine and the other is a human, the random experiment $t_{m,h}$ has only two possible outcomes: $H(m) \wedge M(h)$ or $M(m) \wedge H(h)$. In other words, the sample space of $t_{m,h}$ is

$$\Omega_{m,h} \coloneqq \{H(m) \wedge M(h), M(m) \wedge H(h)\}$$

---

[9] Two events $E_1$ and $E_2$, belonging to the sigma-algebras $\Sigma(\Omega_1)$ and $\Sigma(\Omega_2)$, respectively, are *equivalent*: $\Leftrightarrow$ ($E_1$ occurs $\Leftrightarrow E_2$ occurs). Let $\mathbf{p}_1$ and $\mathbf{p}_2$ be the probability functions defined on $\Sigma(\Omega_1)$ and $\Sigma(\Omega_2)$, respectively. The *Principle of Equivalence* (Giunti et al. 2024, p. 26) states: If $E_1$ is equivalent to $E_2$, then $\mathbf{p}_1(E_1) = \mathbf{p}_2(E_2)$.

Since the sample space is discrete (in fact, finite), the associated sigma-algebra is its powerset $\mathcal{P}(\Omega_{m,h})$. The two events we are interested in are incorrect and correct identification. They are elementary events in $\mathcal{P}(\Omega_{m,h})$:

$$\text{incorrect identification} = F_{m,h} \coloneqq \{H(m) \wedge M(h)\} \in \mathcal{P}(\Omega_{m,h})$$
$$\text{correct identification} = S_{m,h} \coloneqq \{M(m) \wedge H(h)\} \in \mathcal{P}(\Omega_{m,h})$$

Recall that, for any random experiment $t$ with sample space $\Omega$ and sigma-algebra $\Sigma(\Omega)$, an event $E \in \Sigma(\Omega)$ is said to *occur* $:\Leftrightarrow r(t) \in E$, where $r(t) \in \Omega$ is the actual outcome of $t$.

We denote the actual outcome of $t_{m,h}$ by $r(t_{m,h}) \in \mathcal{P}(\Omega_{m,h})$. We further denote the probability function defined on $\mathcal{P}(\Omega_{m,h})$ by $\mathbf{p}_{m,h}$. Let $t_m$ and $t_h$ be the two logically correlated Bernoulli experiments defined in Sec. 1 of this Appendix. We now note that, because of the way the three-player test is constructed, we can establish a priori: $r(t_{m,h}) = H(m) \wedge M(h) \Leftrightarrow r(t_m) = H(m) \Leftrightarrow r(t_h) = M(h)$ and $r(t_{m,h}) = M(m) \wedge H(h) \Leftrightarrow r(t_m) = M(m) \Leftrightarrow r(t_h) = H(h)$. Therefore, $F_{m,h}$ occurs $\Leftrightarrow$ $S_m$ occurs $\Leftrightarrow$ $F_h$ occurs and $S_{m,h}$ occurs $\Leftrightarrow$ $F_m$ occurs $\Leftrightarrow$ $S_h$ occurs. Therefore, by the principle of equivalence, it follows that $\mathbf{p}_{m,h}(F_{m,h}) = \mathbf{p}_m(S_m) = \mathbf{p}_h(F_h)$ and $\mathbf{p}_{m,h}(S_{m,h}) = \mathbf{p}_m(F_m) = \mathbf{p}_h(S_h)$.

### *4. Note on the 2014 Turing Test at the Royal Society (Eugene Goostman)*

Warwick and Shaw (2016) argued that the 30% threshold of incorrect identifications was reached by Eugene Goostman, a program that was identified as human 10 times out of 30 (33%) in a Turing Test conducted on June 6–7, 2014 at the Royal Society in London. In fact, Warwick and Shaw's thesis is not justified. The 30% threshold was set by Turing for a three-player test—that is, a test in which (a) the interrogator converses with two interlocutors, one a machine and the other a human, and (b) can provide only two responses: either "x is a machine and y is a human" or "x is a human and y is a machine" (Turing 1950, p. 433). However, as can be seen from the report (Warwick and Shaw 2016, p. 993), the test satisfied condition (a) but not (b), since also the responses "x is a machine and y is a machine" and "x is a human and y is a human" were allowed. Therefore, despite appearances, that test was not a three-player test. From a formal point of view, it is instead a particular type of two-player test because, like the two-player test, it consists of two Bernoulli experiments that are not logically correlated (see Sec. 2 of this Appendix). What differentiates it from the standard form of the two-player test is only the non-essential fact that the two Bernoulli experiments are conducted in parallel rather than on separate runs. The distinction between three-player and two-player tests is important, because the 30% threshold of incorrect identifications of the machine does not have the same significance in the two types of tests. In fact, only in the three-player test does it represent 6/10 of 50%, i.e., 6/10 of the optimal performance threshold. In the two-player test, on the other hand, it is the percentage of correct identifications of the human that provides an estimate of the optimal performance threshold. Warwick and Shaw do not report this percentage, so it is not possible to estimate such threshold and, consequently, whether 33% is greater than 6/10 of the latter.

## References

Bayne T., Williams, I. (2023). “The Turing test is not a good benchmark for thought in LLMs”, in *Nature Human Behavior*, 7, pp. 1806–1807, <https://doi.org/10.1038/s41562-023-01710-w>.

Biever C. (2023). “ChatGPT broke the Turing Test: The race is on for new ways to assess AI”, in *Nature*, *619*, pp. 686–689, <https://doi.org/10.1038/d41586-023-02361-7>.

French R. M. (2000), “The Turing test: The first fifty years”, in *Trends in Cognitive Sciences*, 4(3), pp. 115-121, <https://www.researchgate.net/publication/222709318>.

Giunti M., Pinna S., Garavaglia F. G. (2024), “Multiple systems in probability theory applications”, in *Multiple Systems*, Contributions to Management Science, Minati G. and Penna M. P. (eds.), Springer Nature Switzerland AG,
<https://link.springer.com/chapter/10.1007/978-3-031-44685-6_2>.

Gonçalves B. (2022), “The Turing test is a thought experiment”, in *Minds and Machines* <https://doi.org/10.1007/s11023-022-09616-8>.

Jannai D., Meron A., Lenz B., Levine Y., Shoham Y. (2023), “Human or not? A gamified approach to the Turing test”, arXiv:2305.20010v1, <https://doi.org/10.48550/arXiv.2305.20010>.

Jones C. R., Bergen B. K. (2024a), “Does GPT-4 pass the Turing test?”, in *Proceedings of the 2024 Conference of the North American Chapter of the Association for Computational Linguistics: Human Language Technologies (Volume 1: Long Papers)*, pp. 5183–5210,
<https://aclanthology.org/2024.naacl-long.290.pdf>.

Jones C. R., Bergen B. K. (2024b), “People cannot distinguish GPT-4 from a human in a Turing test”, arXiv:2405.08007v1, <https://doi.org/10.48550/arXiv.2405.08007>.

Jones C. R., Bergen B. K. (2025), “Large Language Models pass the Turing test”, arXiv:2503.2374v1, <https://doi.org/10.48550/arXiv.2503.23674>.

Jones C. R., Rathi I., Taylor S., and Bergen B. K. (2025). “People cannot distinguish GPT-4 from a human in a Turing test”. In *The 2025 ACM Conference on Fairness, Accountability, and Transparency (FAccT ’25)*, June 23–26, 2025, Athens, Greece. ACM, New York, NY, USA, pp. 1615-1639, <https://doi.org/10.1145/3715275.3732108>.

Oppy G., Dowe D. (2021), “The Turing test”, in *The Stanford Encyclopedia of Philosophy* (Winter 2021 Edition), Zalta E. N. (ed.),
<https://plato.stanford.edu/archives/win2021/entries/turing-test/>.

Restrepo Echavarría R. (2025), “ChatGPT-4 in the Turing test”, in *Minds and Machines*, 35(8), <https://doi.org/10.1007/s11023-025-09711-6>.

Saygin A., Cicekli I., Akman V. (2000), “Turing test: 50 years later”, in *Minds and Machines*, 10(4), pp. 463-518, <https://link.springer.com/article/10.1023/A:1011288000451>.

Sterne T. E. (1954), “Some remarks on confidence or fiducial Limits”, in *Biometrika* , 41(1-2), pp. 275-278, <https://doi.org/10.2307/2333026>.

Turing A. (1950), “Computing machinery and intelligence”, in *Mind*, 59(236), pp. 433-460, <https://doi.org/10.1093/mind/LIX.236.433>.

Warwick K., Shah H. (2016), “Can machines think? A report on Turing test experiments at the Royal Society”, in *Journal of Experimental & Theoretical Artificial Intelligence*, 28(6), pp. 989-1007, <https://doi.org/10.1080/0952813X.2015.1055826>.